\documentclass[11pt]{article}
\usepackage[preprint]{acl}

\usepackage{times}
\usepackage{latexsym}
\usepackage[T1]{fontenc}
\usepackage[utf8]{inputenc}
\usepackage{microtype}
\usepackage{inconsolata}
\usepackage{graphicx}
\usepackage{booktabs}
\usepackage{amsmath}
\usepackage{eurosym} 
\usepackage{xurl} 
\usepackage{longtable} 

\title{From Content to Audience: A Deployed Multimodal Annotation\\
Framework for Broadcast Television Analytics}

\author{Paolo Cupini \\ Politecnico di Milano \\ \texttt{paolo.cupini@mail.polimi.it} \\
  \And Francesco Pierri \\ Politecnico di Milano \\ \texttt{francesco.pierri@polimi.it}}

\begin{document}
\maketitle

\begin{abstract}
Broadcast audience measurement tracks \emph{how many} viewers are watching, but not \emph{why} engagement changes.
We present a deployed multimodal annotation system, developed with a national broadcast media group, that assigns minute-level semantic labels to television news content and links them to disaggregated audience data.
The system combines visual input, ASR, speaker diarization, episode metadata, and an MLLM annotator across four tasks: topic classification, environment classification, visual named-entity recognition, and sensitive-content detection.
Using a 100-clip diagnostic benchmark, we compare configurations under quality, latency, and token-cost constraints, finding that transcripts and metadata drive language-oriented tasks, while environment and sensitive-content labels remain primarily vision-driven.
Native video does not consistently outperform sampled frames; its gains are limited to specific model--task pairs, and smaller models can degrade under extended multimodal context.
We deploy the selected configuration---Gemini-3-Pro with video and metadata---on 14 full episodes (3{,}104 minutes), integrate the resulting annotations with normalized age-cohort ratings, and use them for observational audience-sensitivity analysis and large-scale guest-composition auditing.
We conclude with practical lessons for moving multimodal annotation pipelines into production.
\end{abstract}

\section{Introduction}
\label{sec:introduction}

Broadcast television is monitored through minute-level audience measurement
systems that report metrics such as Average Minute Rating (AMR), capturing how
many viewers are watching at each moment and their demographic
composition~\citep{auditel_methodology_en,tamireland_understanding_tv_data_2025}.
These indicators quantify engagement but do not explain it: connecting
minute-level audience variation to the semantic structure of the underlying
content is still largely manual and does not scale~\citep{gambaro2021_hard_soft_news,hinami2016_audience_behavior_mining}.

Multimodal large language models (MLLMs) can jointly reason over visual frames,
speech transcripts, and structured metadata, and are increasingly used for
automated video annotation~\citep{yin2024_mllm_survey,tang2023_vidllm_survey}.
In principle they could label broadcast programs at minute resolution and align
those labels with audience data. In practice, deployment introduces constraints
that rarely surface in lab settings: content is temporally structured and
heterogeneous, ASR introduces noise that propagates
downstream~\citep{li2024_asr_errors_downstream}, large-scale annotation must
remain computationally sustainable, and it is not obvious that richer inputs
(e.g., full video) justify their latency and token
cost~\citep{qian2024_videostreaming,jiang2025_storm,brkic2025_frame_sampling}.

This paper reports a real-world case study conducted with a national broadcast
media group. We build a multimodal semantic annotation system for professional
TV audience analysis and validate it on more than 3{,}000 minutes of broadcast
content integrated with disaggregated audience measurement data. Two questions
frame the work from a deployment standpoint:
\begin{itemize}\itemsep2pt
  \item \textbf{RQ1.} Which pipeline configuration is production-viable for
  broadcast annotation when task quality is weighed jointly with computational
  cost (tokens, latency)?
  \item \textbf{RQ2.} Can the resulting annotations be \emph{operationally
  integrated} with minute-level audience data---and with content audits such as
  on-screen guest composition---to enable demographic engagement analysis at
  broadcast scale?
\end{itemize}

We proceed in two stages. First, a diagnostic benchmark of 100 annotated clips
is used to compare frame-based and video-based configurations across nine
MLLMs and four tasks, explicitly accounting for token consumption, in order to
select a single production configuration (Section~\ref{sec:config}). Second,
the selected pipeline is deployed on 14 full episodes aired in 2025
(3{,}104 minutes); minute-level annotations are aligned with normalized AMR
disaggregated by age cohort (Section~\ref{sec:deployment}). Our central
practical result is that, for the language-oriented tasks (topic and entity
recognition), speech- and text-derived context (ASR plus metadata) is a more
robust performance driver than temporal visual
continuity~\citep{brkic2025_frame_sampling,cores2025_lost_in_time}, with direct
consequences for how such systems should be built. We close with the lessons
learned (Section~\ref{sec:lessons}).

\section{Related Work}
\label{sec:related_work}

Broadcast programming differs from short-form web video: stable production
formats, recurring guest ecosystems, and strong coupling between spoken
discourse and visual context~\citep{hong2021faces_tv_news}. Prior work mostly
optimizes individual components---speaker recognition from face and audio
embeddings~\citep{chung2018voxceleb2}, scene-level environment labels, and
ASR-driven topic analysis~\citep{radford2023whisper}---rather than end-to-end
pipelines under heterogeneous tasks and constrained budgets. Vision--language
models offer a unified framework for visual and textual
reasoning~\citep{qwen_vl_2023,tang2023_vidllm_survey,alayrac2022_flamingo,liu2023_llava},
but show persistent limits in fine-grained spatiotemporal reasoning: extending
the context window is insufficient without careful input structuring, and frame sampling strongly shapes outcomes~\citep{qian2024_videostreaming,jiang2025_storm,brkic2025_frame_sampling}. Methodologically,
single automatic metrics can mislead system-level comparison and are better
treated as diagnostics~\citep{deutsch-etal-2022-limitations}, performance is
highly sensitive to prompt and input formulation~\citep{gonen-etal-2023-demystifying},
and domain-specific benchmarks must be purpose-built~\citep{chen-etal-2021-finqa}.

A third strand links audiovisual content to audience behavior:
\citet{hong2021faces_tv_news} show that large-scale visual analysis of TV news
reveals systematic editorial exposure patterns, and
\citet{vermeer2025whos_watching} link automated subtitle analysis to
segment-level audience dynamics. Both remain unimodal and depend on existing
subtitles or curated metadata; fully automated multimodal pipelines that
annotate raw broadcast video and integrate it with minute-level, disaggregated
audience data remain underexplored---the gap this work addresses.

\begin{figure*}[!t]
  \centering
\includegraphics[width=.9\linewidth]{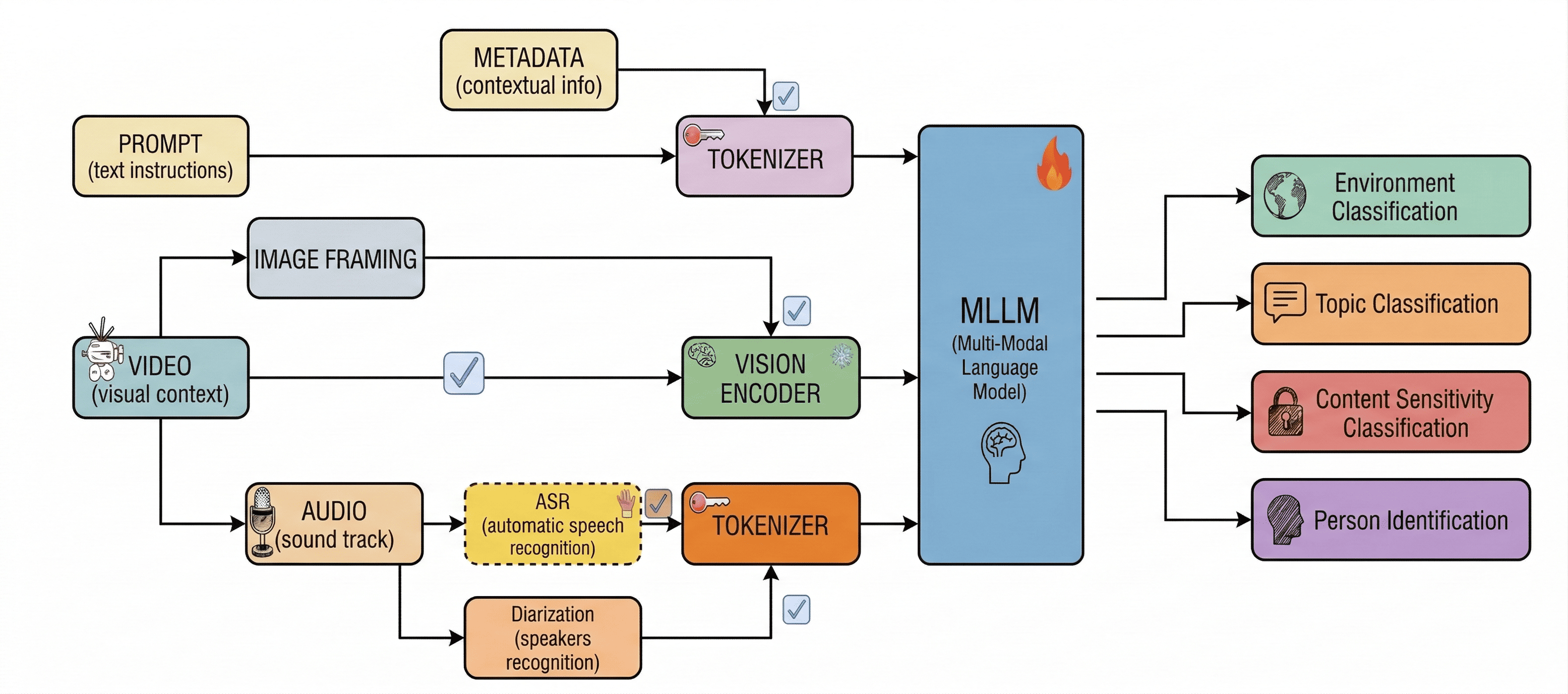}
  \caption{Unified pipeline architecture. Dashed components are optional; the
  visual branch is instantiated either as sampled static frames (frame
  pipeline) or as the full video clip (video pipeline). The audio--textual
  branch is identical across both configurations.}
  \label{fig:pipeline}
\end{figure*}

\section{Methodology}
\label{sec:system}

\noindent\textbf{Pipeline.}
The production system (Figure~\ref{fig:pipeline}) shares a single architecture
across all configurations. An audio--textual branch transcribes speech with
Faster-Whisper, optionally adds speaker diarization via \texttt{pyannote}, and
optionally appends episode-level metadata (program title, broadcast date,
genre, and expected guests). A visual branch is instantiated either as sampled
static frames or as the native video clip. The branches feed an MLLM that
returns, per minute, a structured JSON record over four tasks. Activating
subsets of the optional components yields a grid of input configurations,
ranging from visual-only to visual$+$ASR$+$diarization$+$metadata, that lets us
isolate the contribution of each textual source. The exact configuration set
differs slightly between the two visual branches and is enumerated in
Appendix~\ref{app:fulltables}: notably, the video branch includes a
\emph{visual$+$metadata} configuration without ASR---the one ultimately
selected for deployment (Section~\ref{sec:deployment}).

\noindent\textbf{Tasks and taxonomy.}
Each minute is annotated along four dimensions, inspired by the IAB Content
Taxonomy~\citep{iab_content_taxonomy} and adapted to Italian prime-time
television. \emph{Topic classification} assigns one label from a 28-label
taxonomy (14 macro-areas). \emph{Environment classification} assigns one of 22
setting labels. \emph{Visual named-entity recognition} lists individuals
visually present (empty if none). \emph{Sensitive-content detection} flags
material across six categories drawn from content-moderation and broadcast
regulatory frameworks. The taxonomy was refined iteratively with broadcast
domain experts; the full label set is in Appendix~\ref{app:taxonomy}.

\noindent\textbf{Deployment setting.}
The selected configuration (Gemini-3-Pro with video and episode-level
metadata) is deployed on 14 full episodes of a single
fixed-slot weekly program, annotating 3{,}104 minutes at one-minute
granularity (advertising excluded). Audience metrics are supplied by an Italian
media company in aggregated, normalized form; no personally identifiable
information is accessed.

\section{Choosing the Production Configuration}
\label{sec:config}

\noindent\textbf{Diagnostic benchmark.}
To select a configuration under realistic constraints, we built a diagnostic
test set of 100 one-minute clips from Italian broadcast programs aired in 2025,
stratified across four editorial macro-categories---current-affairs talk shows,
investigative, cultural, and lifestyle programs---with 25 clips sampled per
category (one per episode) to balance editorial conditions by design; program
identities are anonymized. Clips are extracted without enforcing scene
boundaries and may span multiple shots, speaker turns, and topic transitions,
reflecting operational conditions rather than curated single-topic segments. Two annotators with TV-content experience labeled all
clips; inter-annotator agreement (Cohen's $\kappa$) was 0.70 (topic), 0.67
(environment), and 0.67 (entities); sensitive content reached $\kappa{=}1.0$,
interpreted cautiously given only 12 positive clips. Disagreements were
resolved by consensus. We evaluate nine MLLMs spanning proprietary and
open-weight families---Gemini-3-Pro, GPT-5-mini, Llama-4-Maverick,
Qwen2.5-VL-72B, Qwen3-VL (235B/30B/8B), Gemma-3-27B, and Ministral-14B---in the
frame setting, and the six models supporting native video input (Gemini-3-Pro,
Qwen3-Omni, Qwen3-VL-235B, Qwen3-VL-30B, Molmo-2-8B, and Seed-1.6) for the
cross-pipeline comparison, in a zero-shot setting with a single prompt template
that includes the full taxonomy (Appendix~\ref{app:prompts}).
Predictions outside the label set are scored as incorrect. Input token counts
from model APIs are reported as an explicit proxy for cost.

\noindent\textbf{Component choices.}
Preliminary studies fixed the visual and ASR components. Accuracy plateaus
beyond 12 frames across all models, and shot-based sampling offers no advantage
on this material; uniform sampling at 0.2~fps (12 frames) cuts token use by
${\sim}40\%$ relative to an 18-frame budget at comparable accuracy. Across the
ASR systems compared, Faster-Whisper Medium offers competitive downstream
quality without the latency overhead of larger variants, making it suitable for
cost-efficient large-scale inference. This fixes the ASR component for the
benchmark configurations that include it; the configuration ultimately deployed
omits ASR altogether (Section~\ref{sec:deployment}).

\noindent\textbf{What drives quality.}
Task quality is shaped by the interaction of input modality and model capacity,
not by raw visual richness (Table~\ref{tab:summary}). The frame pipeline's
\texttt{only}/\texttt{asr}/\texttt{asr\_meta} configurations let us isolate each
textual source within a fixed model: \texttt{only}$\rightarrow$\texttt{asr}
isolates the ASR increment, \texttt{asr}$\rightarrow$\texttt{asr\_meta} the
metadata increment. For \emph{topic}, the isolated ASR increment is
large and consistent ($+0.14$ to $+0.21$ accuracy for Gemini-3-Pro,
Llama-4-Maverick, and the Qwen models), while the subsequent metadata increment
is near zero: speech-derived text is the driver and diarization adds no
systematic benefit. For \emph{visual NER}---the most demanding task---the
pattern reverses: the ASR increment is negligible whereas the isolated metadata
increment dominates (precision $0.08\rightarrow0.52$ for Qwen3-VL-235B and
Qwen3-VL-30B, $0.42\rightarrow0.60$ for Gemini-3-Pro), though enrichment is not
uniformly safe---diarization noise degrades identity resolution for several
models, and smaller models hallucinate identities without textual grounding.
\emph{Environment} is primarily perceptual; whether native video helps is
model-dependent (Table~\ref{tab:controlled}), with gains concentrated in
motion-dependent or spatially ambiguous scenes.
\emph{Sensitive-content} detection is dominated by false-positive control:
transcripts can contextualize ambiguous scenes but also trigger spurious
activations via lexical shortcuts, especially in smaller models; with only 12
positive clips, its metrics are diagnostic and do not drive
configuration selection.

\noindent\textbf{Video is not universally better (controlled comparison).}
Critically, native video input does not systematically outperform sampled
frames. To isolate the visual representation from confounds of model capacity
and textual input, we compare the two pipelines \emph{within} each of the three
architectures common to both (Gemini-3-Pro, Qwen3-VL-235B, Qwen3-VL-30B) at
matched input compositions (Table~\ref{tab:controlled}). Holding model and
textual inputs fixed, the video$-$frame accuracy gap is small and, in most cells, not statistically
distinguishable from zero: of the eighteen matched comparisons, twelve have a
95\% bootstrap confidence interval (paired over clips) that spans zero. For
\emph{topic}, video helps only when no transcript is present (Gemini-3-Pro
$+0.12$, CI~$[{+}0.05,{+}0.20]$; Qwen3-VL-30B $+0.14$, CI~$[{+}0.05,{+}0.24]$)
and the gain becomes indistinguishable from zero once ASR is added. For
\emph{environment}, the significant effects are model-specific and of opposite
sign: video helps Qwen3-VL-30B substantially ($+0.36$, CI~$[{+}0.23,{+}0.48]$)
but \emph{hurts} the larger Gemini-3-Pro ($-0.10$, CI~$[{-}0.19,{-}0.01]$). The
residual token gap between the two encodings is intrinsic to the representation
and is reported, not equalized (Appendix~\ref{app:fulltables}). This
within-model evidence---bounded to three architectures and a 100-clip
set---substantiates the headline result under controlled conditions: the video
benefit does not track model size and is confined to a few model--task cells,
while degradation under extended multimodal context appears among the smaller
video-only models in the full results (Appendix~\ref{app:fulltables}). Combined
with the token-cost accounting, this argues against adopting video ingestion as
a default. We select Gemini-3-Pro with video plus metadata for deployment: it is
not Pareto-dominant (accepting a small environment loss, 0.73 vs.\ 0.76 for
frames$+$ASR$+$metadata) but gives the best topic accuracy (0.83) at roughly
half the token cost. It omits ASR---for the highest-capacity model, metadata
already supplies enough textual context (topic 0.83 vs.\ 0.80 with ASR)---which
in a corporate setting also drops a transcription stage and its hardware,
lowering latency and improving portability.

\begin{table*}[t]
\centering\small
\setlength{\tabcolsep}{6pt}
\renewcommand{\arraystretch}{1.05}
\begin{tabular}{@{}ll rl rl@{}}
\toprule
\textbf{Model} & \textbf{Input} & \textbf{$\Delta$Topic} & \textbf{95\% CI} & \textbf{$\Delta$Env} & \textbf{95\% CI} \\
\midrule
Gemini-3-Pro   & only     & $\mathbf{+0.12}$ & $[{+}0.05,{+}0.20]$ & $-0.04$ & $[{-}0.13,{+}0.05]$ \\
               & asr+meta & $+0.01$ & $[{-}0.04,{+}0.06]$ & $\mathbf{-0.10}$ & $[{-}0.19,{-}0.01]$ \\
               & +diar    & $+0.01$ & $[{-}0.03,{+}0.05]$ & $+0.01$ & $[{-}0.08,{+}0.10]$ \\
\addlinespace
Qwen3-VL-235B  & only     & $+0.04$ & $[{-}0.04,{+}0.12]$ & $+0.06$ & $[{-}0.02,{+}0.14]$ \\
               & asr+meta & $+0.01$ & $[{-}0.07,{+}0.10]$ & $+0.04$ & $[{-}0.05,{+}0.13]$ \\
               & +diar    & $+0.05$ & $[{-}0.06,{+}0.15]$ & $+0.03$ & $[{-}0.07,{+}0.13]$ \\
\addlinespace
Qwen3-VL-30B   & only     & $\mathbf{+0.14}$ & $[{+}0.05,{+}0.24]$ & $\mathbf{+0.19}$ & $[{+}0.09,{+}0.30]$ \\
               & asr+meta & $+0.05$ & $[{-}0.04,{+}0.14]$ & $\mathbf{+0.36}$ & $[{+}0.23,{+}0.48]$ \\
               & +diar    & $+0.07$ & $[{-}0.02,{+}0.16]$ & $\mathbf{+0.20}$ & $[{+}0.09,{+}0.32]$ \\
\bottomrule
\end{tabular}
\caption{Controlled within-model comparison: accuracy gap
$\Delta=\text{video}-\text{frame}$ for the three architectures present in both
pipelines, at matched input compositions (\texttt{only}, \texttt{asr+meta},
\texttt{+diar}$=$\texttt{asr+diar+meta}). Deltas are computed paired over clips
with a prediction in both pipelines ($n{=}98$--$100$); 95\% CIs are
percentile bootstrap intervals (10K resamples). \textbf{Bold} marks intervals
that exclude zero. Holding model and textual inputs fixed, twelve of eighteen
comparisons are statistical ties and the few significant effects are
model-specific and of opposite sign---video does not systematically outperform
frames. Absolute values are in Appendix~\ref{app:fulltables}.}
\label{tab:controlled}
\end{table*}

\noindent\textbf{Error patterns.}
Manual inspection reveals failure modes the aggregate metrics obscure. Topic
errors are systematic collapses between adjacent labels
(\emph{international}\,$\rightarrow$\,\emph{domestic politics}), driven by
shared lexical fields in the transcript; environment errors cluster within
perceptually similar settings (\emph{home apartment} vs.\ \emph{corporate
office}), while distinctive studio layouts stay stable. Sensitive-content
errors stem from threshold instability---under weak visual evidence, lexical
triggers induce spurious activations, turning text into a liability---and
entity-recognition failures concentrate in smaller models and in clips lacking
textual anchors, where models hallucinate identities. Model capacity moderates
these effects but does not remove the underlying bottlenecks.

\begin{table}[t]
\centering\small
\setlength{\tabcolsep}{4pt}
\renewcommand{\arraystretch}{1.05}
\begin{tabular}{@{}llcl@{}}
\toprule
\textbf{Task} & \textbf{Driver} & \textbf{Video} & \textbf{Failure mode} \\
\midrule
Topic       & Speech   & Low  & Semantic overlap \\
Environment & Visual   & Var. & Spatial ambiguity \\
Sensitive   & Visual   & n/a  & Spurious activation \\
Entities    & Metadata & Low  & Identity halluc. \\
\bottomrule
\end{tabular}
\caption{Dominant modality and characteristic failure mode per task, relative
to the frame baseline. ``Video'' is the benefit of native video over sampled
frames, as established by the controlled comparison (Table~\ref{tab:controlled}):
\emph{Low}$=$gains only without a transcript or not significant;
\emph{Var.}$=$model-dependent, significant for one model and reversed for
another; \emph{n/a}$=$diagnostic task, 12 positives. Full per-model tables are
in Appendix~\ref{app:fulltables}.}
\label{tab:summary}
\end{table}

\section{Deployment and Audience Analytics}
\label{sec:deployment}

\noindent\textbf{Production cost and throughput.}
Table~\ref{tab:cost} summarizes deployment economics. The selected
configuration consumes ${\sim}6.4$K input tokens per annotated minute and emits
only four structured fields, so annotating the full corpus costs
${\approx}\$40$ at list pricing\footnote{Gemini~3~Pro list pricing of \$2.00 per
1M input tokens for prompts ${\le}200$K tokens; output negligible.
\url{https://ai.google.dev/gemini-api/docs/pricing}, accessed June 2026. The
asynchronous Batch API halves these rates.} (${\approx}\$20$ via the Batch
API)---under a cent and a half per minute. Median latency (${\sim}31$ s/minute,
real-time factor ${\approx}0.5$) means a single request stream keeps pace with
broadcast; annotation is offline and trivially parallelizable. Notably, native
video$+$metadata is also \emph{more} token-economical than the frame
configuration (${\sim}6.4$K vs ${\sim}14.6$K tokens/minute): the model encodes
native video at a low fixed token rate per sampled second, whereas twelve
full-resolution stills each consume a large token block. The production choice
is therefore simultaneously the higher-quality and the cheaper one.

\begin{table}[t]
\centering\small
\setlength{\tabcolsep}{6pt}
\renewcommand{\arraystretch}{1.05}
\begin{tabular}{@{}lr@{}}
\toprule
\textbf{Production metric} & \textbf{Value} \\
\midrule
Input tokens / annotated minute        & ${\sim}6{,}443$ \\
Total input tokens (3{,}104 min)       & ${\sim}20.0$M \\
Output tokens / call                   & 4 fields (negl.) \\
API latency / minute                   & ${\sim}31$ s \\
Real-time factor                       & ${\approx}0.52$ \\
Est.\ cost, list (\$2 / 1M in)         & ${\approx}\$40$ \\
\quad per annotated minute             & ${\approx}\$0.013$ \\
\quad per hour of broadcast            & ${\approx}\$0.8$ \\
Est.\ cost, Batch API ($-50\%$)        & ${\approx}\$20$ \\
\bottomrule
\end{tabular}
\caption{Production cost and throughput of the deployed configuration
(Gemini-3-Pro, video$+$metadata) over the 3{,}104-minute corpus. Costs use
Gemini~3~Pro list pricing (\$2.00 per 1M input tokens, ${\le}200$K context);
output is limited to the four structured fields and is negligible.}
\label{tab:cost}
\end{table}
\noindent\textbf{Human-cost comparison.}
The media partner's manual workflow for the same deliverable is estimated at
roughly two analyst-hours per episode (one to annotate content, one to ingest
audience data and produce the analysis): ${\approx}28$ hours per 14-episode
cycle, or ${\approx}\,$\euro{}840 at an illustrative \euro{}30/h, against
${\approx}\$40$ of compute plus the residual human time to confirm
sensitive-content flags---an order-of-magnitude gap that widens with scale,
since manual effort grows linearly with airtime while per-minute annotation
cost is fixed. The manual figure reflects a coarse operational pass, not the
research-grade annotation used for the benchmark, and only bounds the saving.

\noindent\textbf{Content characterization.}
On the 3{,}104 deployed minutes, editorial structure is highly concentrated:
\emph{domestic politics} (40.8\%) and \emph{international politics} (21.7\%)
account for over 60\% of airtime, followed by \emph{economy and finance}
(11.7\%), \emph{humor and satire} (9.2\%), and \emph{crime and justice}
(6.0\%); remaining topics form a long tail. The NER module identifies 143
unique guests (hosts excluded) over 3{,}717 minute-level occurrences.
Participation is structurally asymmetric: male guests account for 81.1\% of
occurrences (89 individuals, 33.9 appearances each on average) versus 700
occurrences across 54 female guests (13.0 each), and 80.2\% of minutes feature
exclusively male guests---a pattern replicated across episodes. Since visual
NER operates at F1$\,{\approx}\,$0.60 at deployment, these absolute
participation figures should be read as approximate estimates rather than exact
counts; the asymmetry itself, however, is large and directionally robust to this
error rate, illustrating the kind of large-scale content audit the pipeline
makes feasible.

\begin{figure*}[!t]
  \centering
  \includegraphics[width=0.6\textwidth]{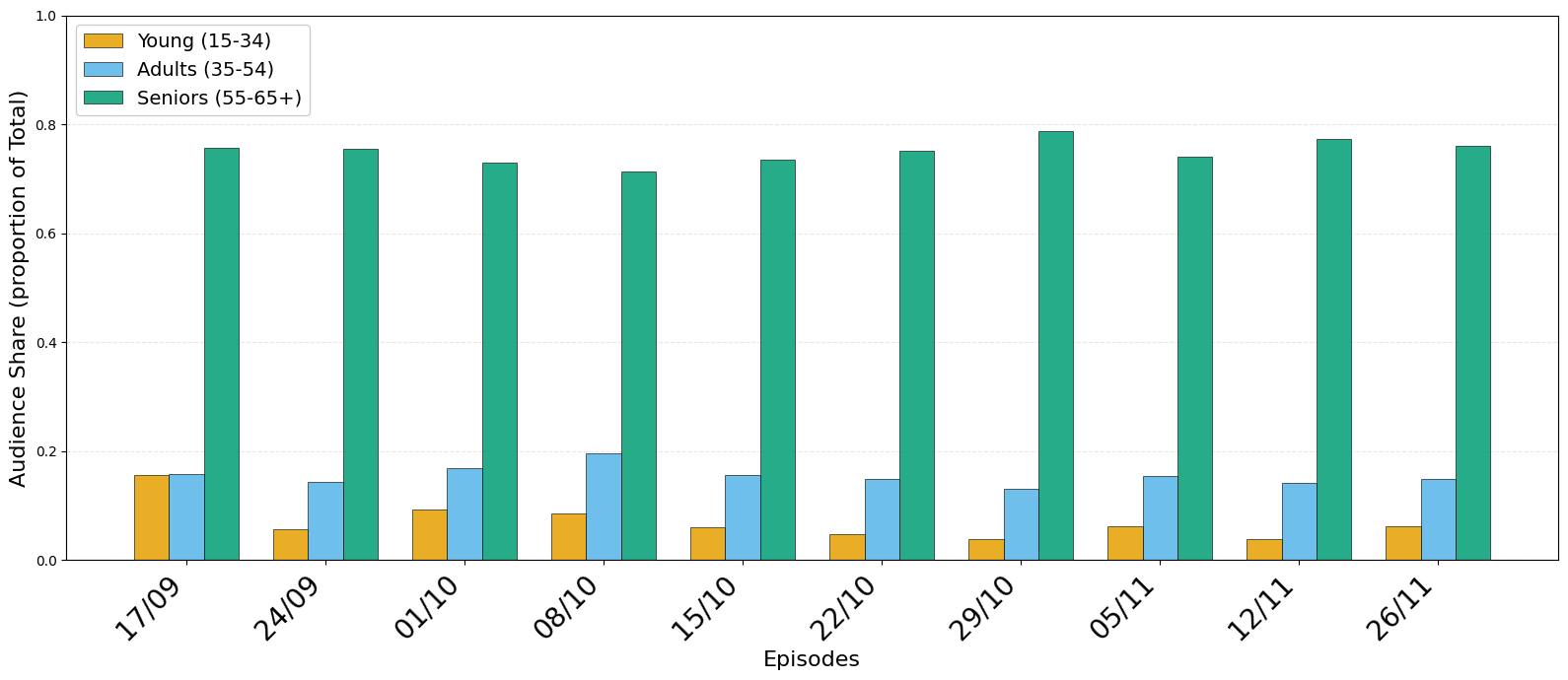}
  \caption{Normalized AMR per episode by age cohort. Seniors (55+) are the most
  stable segment; younger viewers vary more across episodes.}
  \label{fig:audience}
\end{figure*}

\noindent\textbf{Baseline audience composition.}
Figure~\ref{fig:audience} reports normalized AMR per episode by age cohort
(Young 15--34, Adults 35--54, Seniors 55+). The senior segment is the largest
and most stable component; younger viewers show markedly higher inter-episode
variability, indicating that episode-level fluctuations are driven mainly by
changes in younger participation rather than the senior base.

\noindent\textbf{Topic-level sensitivity and divergence.}
Audience values are normalized intra-episode via per-cohort $z$-scores
(Appendix~\ref{app:zscore}), so positive values indicate above-average
engagement controlling for episode
popularity and seasonality. Among the ten topics with the largest inter-cohort
$z$-score gap, \emph{art and literature} is positive
for Young viewers yet negative for Adults and Seniors---the clearest case of
cohort-selective engagement. A gradient pattern characterizes \emph{music},
\emph{health and wellness}, and \emph{sports--football}: negative across
cohorts, with the decline steepening with age. \emph{Family and relationships}
and \emph{environment and climate} are associated with disengagement across all
groups, while \emph{food and cooking} and \emph{sports--other} act as shared
attractors that differ in amplitude rather than direction.

\noindent\textbf{Interpretation.}
High-frequency structural topics sustain a stable engagement backbone, while
low-frequency episodic segments generate localized deviations that
differentially affect age cohorts---a shift from static aggregate measurement
toward dynamic audience-sensitivity analysis of direct value for editorial
planning. We present this integration as an illustrative downstream use enabled
by the annotations rather than a standalone analytical contribution; all
reported patterns are observational and no causal claims are made.

\section{Lessons Learned and Conclusion}
\label{sec:lessons}

\noindent\textbf{Speech and text beat video for the language-oriented tasks.}
Video pipelines do not systematically outperform frame configurations: under a
controlled within-model comparison, most matched gaps are statistical ties and
the few significant effects are model-specific and of opposite sign, not a
function of model size (degradation under extended multimodal context shows up
mainly in the smaller video-only models). Video ingestion should be a
conditional design choice, validated per task and per model before its cost is
accepted.

\noindent\textbf{Metadata is the cheapest high-impact lever.}
Isolating each textual increment, episode-level metadata was the dominant driver
for \emph{entity resolution}---outweighing transcription quality there at
negligible token cost---whereas speech transcripts, not metadata, drove topic.
In a deployment, cheap structured side information is often more valuable than a
heavier perceptual modality for the tasks it targets.

\noindent\textbf{Taxonomy design is a first-order engineering decision.}
Recurring errors---semantic collapse between adjacent topics, perceptual
confusion between similar environments---were attributable to label structure,
not model capacity: overlapping or underspecified categories introduce
systematic ambiguity that no extra modality resolves. The taxonomy should be a
maintained engineering artifact, not a fixed schema.

\noindent\textbf{Automated annotation enables scale but needs error-aware use.}
Annotating at inference speed makes longitudinal audits and cross-program
comparisons feasible, but systematic label biases (environment
misclassification, topic conflation) propagate downstream. We add two
safeguards: the lowest-precision task (sensitive-content detection) is
human-confirmed before any downstream use rather than driving autonomous
decisions, and a rolling sample of deployed minutes is periodically
re-annotated to monitor drift, seeded by the diagnostic benchmark. Studies
built on automated annotations should carry reliability proxies, and findings
from borderline categories should be read as indicative.


Taken together, these lessons show that deployed multimodal annotation should
match each task to the cheapest reliable signal rather than defaulting to the
richest input. In our broadcast setting, speech supported topic interpretation,
metadata stabilized entity resolution, and vision remained necessary for
perceptual labels. Under these safeguards, such annotations can support scalable
audience analysis and content audits without treating automated labels as
ground truth.

\clearpage

\section*{Limitations}

The diagnostic benchmark (100 clips) limits statistical robustness and
generalizability and should be read as diagnostic rather than as a large-scale
corpus. The cross-pipeline comparison is controlled \emph{within} the three
architectures common to both pipelines and at matched textual inputs, but video
and frame encodings cannot be equalized in token count; the residual token gap
is an intrinsic property of the representation rather than a nuisance variable
we control away. Bootstrap confidence intervals are reported for the controlled
cross-pipeline comparison; the remaining per-model tables report point
estimates only. Although the clips are stratified by editorial macro-category,
they remain internally heterogeneous (multiple shots, speakers, and topic
transitions per minute), so per-signal effects are reported descriptively and
may partly reflect clip composition. The taxonomies are tailored to a specific
broadcast format and may require adaptation elsewhere. The evaluation is
monolingual (Italian) and centered on a single program type, leaving
multilingual content and other genres untested. All experiments use zero-shot
prompting of general-purpose models, without task-specific fine-tuning. The
audience analyses are purely observational: intra-episode per-cohort $z$-score
normalization removes episode-level popularity and seasonality, but the analysis
is not benchmarked against simpler scheduling baselines, reported associations
between content and engagement do not establish causal relationships, and they
may be confounded by concurrent programming. Finally, downstream analyses
inherit the annotation error profile of the pipeline, which is only partially
characterized at this scale.

\section*{Ethical Considerations}

Audience data were provided by the media partner in aggregated and normalized
form, disaggregated only by coarse age cohorts; no personally identifiable
information was accessed or stored. The system performs sensitive-content
detection and visual entity recognition over public broadcast material; we
report these capabilities for analytical and editorial-planning purposes and
caution against punitive or surveillance uses. Automated annotation can encode
and amplify bias: the observed gender asymmetry in guest participation is a
property of the broadcast content, but pipeline errors could distort such
measurements, so demographic findings should be validated before informing
decisions. Program and company identities are anonymized, and proprietary data
are not released. We follow the ACL Code of Ethics.

\bibliography{custom}

\appendix

\section{Complete Annotation Taxonomy}
\label{app:taxonomy}
Table~\ref{tab:taxonomies} lists the full label sets for topic, environment,
and sensitive-content annotation.

\begin{table}[h]
  \centering\footnotesize
  \resizebox{\linewidth}{!}{%
  \begin{tabular}{lll}
    \toprule
    \textbf{Topic} & \textbf{Environment} & \textbf{Sensitive} \\
    \midrule
    Domestic politics            & Studio -- Single host          & Violence                        \\
    International politics       & Studio -- 1-to-1 interview     & Blood                           \\
    Crime and justice            & Studio -- Guest panel          & War / armed conflicts           \\
    Environment and climate      & Studio -- Remote split screen  & Organized crime                 \\
    Society and social phenomena & Studio -- Video segment        & Humanitarian crises             \\
    Family and relationships     & Home -- Apartment              & Self-harm / suicide             \\
    Cinema, TV and entertainment & Home -- Kitchen                &                                 \\
    Music                        & Corporate office               &                                 \\
    Humor and satire             & Commercial/public venue        &                                 \\
    Art and literature           & Spa/wellness center            &                                 \\
    History and archaeology      & Vehicle/transport              &                                 \\
    Religion and spirituality    & Generic urban outdoor          &                                 \\
    Science and technology       & Nature -- Outdoors             &                                 \\
    Education and training       & Nature -- Mountain             &                                 \\
    Food and cooking             & Identified tourist site        &                                 \\
    Fashion and beauty           &                                &                                 \\
    Travel and tourism           &                                &                                 \\
    Health and wellness          &                                &                                 \\
    Motors and vehicles          &                                &                                 \\
    Sports -- Football           &                                &                                 \\
    Sports -- Other              &                                &                                 \\
    \bottomrule
  \end{tabular}}
  \caption{Taxonomies for semantic annotation: topic (left), environment
  (center), and sensitive-content categories (right).}
  \label{tab:taxonomies}
\end{table}

\section{Prompt Strategies}
\label{app:prompts}
Both pipelines share the output schema and hard constraints: select exactly one
label for topic and environment, at most one sensitive-content flag, include a
guest name only if the person is visually recognized or explicitly named, and
return JSON only. Configurations differ in which sources are provided and in
their relative authority. When metadata is added, it is explicitly subordinated
to visual evidence (used only to confirm or correct visually recognized
identities, never to assume presence). When ASR is added, a three-level
hierarchy is established: video is primary for visual elements, ASR for spoken
content, and metadata acts as a correction layer for proper nouns and context.

\section{Audience Normalization}
\label{app:zscore}
Topic-level audience sensitivity (Section~\ref{sec:deployment}) is computed on
per-cohort, intra-episode $z$-scores. Let $c$ index the age cohort, $e$ the
episode, and $t$ the topic. Let $x_{c,e,t}$ be the mean normalized AMR of cohort
$c$ over the minutes labeled with topic $t$ in episode $e$, and let
$\mu_{c,e}$ and $\sigma_{c,e}$ be the mean and standard deviation of cohort
$c$'s per-minute normalized AMR within episode $e$. The $z$-score is
\begin{equation}
z_{c,e,t} \;=\; \frac{x_{c,e,t}-\mu_{c,e}}{\sigma_{c,e}} .
\end{equation}
Normalizing within each episode and cohort removes differences in baseline
popularity and seasonality, so positive (negative) values denote above-average
(below-average) engagement of a cohort with a topic \emph{relative to its own
episode-level mean}. The inter-cohort gap discussed in
Section~\ref{sec:deployment} is the spread of $z_{c,e,t}$ across cohorts,
averaged over episodes.

\onecolumn
\section{Full Per-Model Results}
\label{app:fulltables}
Complete per-model, per-configuration results for all four tasks. We report
Accuracy, Precision, Recall, and F1, together with input tokens (\textbf{Tok})
and average API latency in milliseconds (\textbf{Lat}). The video pipeline
covers the six models supporting native video; the frame pipeline covers all
nine models. Tables~\ref{tab:res_vid_env}--\ref{tab:res_vid_per} report the
video pipeline and Tables~\ref{tab:res_frm_env}--\ref{tab:res_frm_per} the
frame pipeline.
\begin{center}
\footnotesize
\setlength{\tabcolsep}{5pt}
\begin{longtable}{llrrrrrr}
\caption{Environment Recognition --- video pipeline. Input: \texttt{asr\_diar\_meta}=video+ASR+diarization+metadata, \texttt{asr\_meta}=video+ASR+metadata, \texttt{meta}=video+metadata, \texttt{only}=video only. \textbf{Tok}=input tokens; \textbf{Lat}=API latency (ms); \texttt{--}=undefined (no positive predictions).}\label{tab:res_vid_env}\\
\toprule
\textbf{Model} & \textbf{Input} & \textbf{Acc} & \textbf{Prec} & \textbf{Rec} & \textbf{F1} & \textbf{Tok} & \textbf{Lat} \\
\midrule
\endfirsthead
\multicolumn{8}{c}{\tablename~\thetable{} -- continued}\\
\toprule
\textbf{Model} & \textbf{Input} & \textbf{Acc} & \textbf{Prec} & \textbf{Rec} & \textbf{F1} & \textbf{Tok} & \textbf{Lat} \\
\midrule
\endhead
\midrule
\multicolumn{8}{r}{\footnotesize continued on next page}\\
\endfoot
\bottomrule
\endlastfoot
      gemini-3-pro & asr\_diar\_meta & 0.74 & 0.79 & 0.74 & 0.75 & 6856 & 31932 \\
      gemini-3-pro & asr\_meta & 0.66 & 0.76 & 0.66 & 0.68 & 6732 & 29595 \\
      gemini-3-pro & meta & 0.73 & 0.79 & 0.73 & 0.75 & 6443 & 31008 \\
      gemini-3-pro & only & 0.66 & 0.73 & 0.66 & 0.68 & 6224 & 35729 \\
      molmo-2-8b & asr\_diar\_meta & 0.33 & 0.42 & 0.33 & 0.29 & 12197 & 22064 \\
      molmo-2-8b & asr\_meta & 0.14 & 0.06 & 0.14 & 0.06 & 11846 & 26757 \\
      molmo-2-8b & meta & 0.26 & 0.68 & 0.26 & 0.24 & 11608 & 24596 \\
      molmo-2-8b & only & 0.18 & 0.60 & 0.18 & 0.14 & 11361 & 21761 \\
      qwen3-omni & asr\_diar\_meta & 0.54 & 0.49 & 0.54 & 0.49 & 15346 & 9525 \\
      qwen3-omni & asr\_meta & 0.56 & 0.56 & 0.56 & 0.53 & 15207 & 9735 \\
      qwen3-omni & meta & 0.55 & 0.63 & 0.55 & 0.54 & 14861 & 9879 \\
      qwen3-omni & only & 0.55 & 0.55 & 0.55 & 0.53 & 14615 & 9913 \\
      qwen3-235b & asr\_diar\_meta & 0.65 & 0.65 & 0.65 & 0.62 & 14614 & 7178 \\
      qwen3-235b & asr\_meta & 0.71 & 0.73 & 0.71 & 0.71 & 14475 & 7237 \\
      qwen3-235b & meta & 0.70 & 0.72 & 0.70 & 0.69 & 14129 & 7091 \\
      qwen3-235b & only & 0.71 & 0.74 & 0.71 & 0.71 & 13884 & 9023 \\
      qwen3-30b & asr\_diar\_meta & 0.62 & 0.58 & 0.62 & 0.57 & 14614 & 22218 \\
      qwen3-30b & asr\_meta & 0.66 & 0.60 & 0.66 & 0.62 & 14475 & 13720 \\
      qwen3-30b & meta & 0.71 & 0.77 & 0.71 & 0.69 & 14130 & 20742 \\
      qwen3-30b & only & 0.70 & 0.78 & 0.70 & 0.69 & 13884 & 18405 \\
      seed-1.6 & asr\_diar\_meta & 0.42 & 0.48 & 0.42 & 0.40 & 19506 & 40369 \\
      seed-1.6 & asr\_meta & 0.54 & 0.65 & 0.54 & 0.56 & 19361 & 34226 \\
      seed-1.6 & meta & 0.61 & 0.64 & 0.61 & 0.61 & 19036 & 33848 \\
      seed-1.6 & only & 0.58 & 0.56 & 0.58 & 0.56 & 18799 & 57734 \\
\end{longtable}
\end{center}

\begin{center}
\footnotesize
\setlength{\tabcolsep}{5pt}
\begin{longtable}{llrrrrrr}
\caption{Topic Classification --- video pipeline. Input: \texttt{asr\_diar\_meta}=video+ASR+diarization+metadata, \texttt{asr\_meta}=video+ASR+metadata, \texttt{meta}=video+metadata, \texttt{only}=video only. \textbf{Tok}=input tokens; \textbf{Lat}=API latency (ms); \texttt{--}=undefined (no positive predictions).}\label{tab:res_vid_top}\\
\toprule
\textbf{Model} & \textbf{Input} & \textbf{Acc} & \textbf{Prec} & \textbf{Rec} & \textbf{F1} & \textbf{Tok} & \textbf{Lat} \\
\midrule
\endfirsthead
\multicolumn{8}{c}{\tablename~\thetable{} -- continued}\\
\toprule
\textbf{Model} & \textbf{Input} & \textbf{Acc} & \textbf{Prec} & \textbf{Rec} & \textbf{F1} & \textbf{Tok} & \textbf{Lat} \\
\midrule
\endhead
\midrule
\multicolumn{8}{r}{\footnotesize continued on next page}\\
\endfoot
\bottomrule
\endlastfoot
      gemini-3-pro & asr\_diar\_meta & 0.81 & 0.82 & 0.81 & 0.80 & 6856 & 31932 \\
      gemini-3-pro & asr\_meta & 0.80 & 0.82 & 0.80 & 0.79 & 6732 & 29595 \\
      gemini-3-pro & meta & 0.83 & 0.82 & 0.83 & 0.81 & 6443 & 31008 \\
      gemini-3-pro & only & 0.79 & 0.79 & 0.79 & 0.77 & 6224 & 35729 \\
      molmo-2-8b & asr\_diar\_meta & 0.53 & 0.68 & 0.53 & 0.53 & 12197 & 22064 \\
      molmo-2-8b & asr\_meta & 0.60 & 0.73 & 0.60 & 0.62 & 11846 & 26757 \\
      molmo-2-8b & meta & 0.50 & 0.67 & 0.50 & 0.52 & 11608 & 24596 \\
      molmo-2-8b & only & 0.34 & 0.62 & 0.34 & 0.36 & 11361 & 21761 \\
      qwen3-omni & asr\_diar\_meta & 0.76 & 0.75 & 0.76 & 0.74 & 15346 & 9525 \\
      qwen3-omni & asr\_meta & 0.77 & 0.79 & 0.77 & 0.77 & 15207 & 9735 \\
      qwen3-omni & meta & 0.78 & 0.80 & 0.78 & 0.77 & 14861 & 9879 \\
      qwen3-omni & only & 0.78 & 0.82 & 0.78 & 0.78 & 14615 & 9913 \\
      qwen3-235b & asr\_diar\_meta & 0.69 & 0.74 & 0.69 & 0.69 & 14614 & 7178 \\
      qwen3-235b & asr\_meta & 0.70 & 0.73 & 0.70 & 0.69 & 14475 & 7237 \\
      qwen3-235b & meta & 0.63 & 0.70 & 0.63 & 0.64 & 14129 & 7091 \\
      qwen3-235b & only & 0.55 & 0.68 & 0.55 & 0.57 & 13884 & 9023 \\
      qwen3-30b & asr\_diar\_meta & 0.69 & 0.77 & 0.69 & 0.70 & 14614 & 22218 \\
      qwen3-30b & asr\_meta & 0.71 & 0.78 & 0.71 & 0.72 & 14475 & 13720 \\
      qwen3-30b & meta & 0.65 & 0.69 & 0.65 & 0.65 & 14130 & 20742 \\
      qwen3-30b & only & 0.60 & 0.68 & 0.60 & 0.60 & 13884 & 18405 \\
      seed-1.6 & asr\_diar\_meta & 0.75 & 0.79 & 0.75 & 0.76 & 19506 & 40369 \\
      seed-1.6 & asr\_meta & 0.71 & 0.70 & 0.71 & 0.70 & 19361 & 34226 \\
      seed-1.6 & meta & 0.60 & 0.73 & 0.60 & 0.63 & 19036 & 33848 \\
      seed-1.6 & only & 0.45 & 0.68 & 0.45 & 0.50 & 18799 & 57734 \\
\end{longtable}
\end{center}

\begin{center}
\footnotesize
\setlength{\tabcolsep}{5pt}
\begin{longtable}{llrrrrrr}
\caption{Sensitive-Content Detection --- video pipeline. \emph{Metrics are diagnostic only}: with 12 positive clips, precision/recall are highly unstable and are \emph{not} used for configuration selection. Input: \texttt{asr\_diar\_meta}=video+ASR+diarization+metadata, \texttt{asr\_meta}=video+ASR+metadata, \texttt{meta}=video+metadata, \texttt{only}=video only. \textbf{Tok}=input tokens; \textbf{Lat}=API latency (ms); \texttt{--}=undefined (no positive predictions).}\label{tab:res_vid_sen}\\
\toprule
\textbf{Model} & \textbf{Input} & \textbf{Acc} & \textbf{Prec} & \textbf{Rec} & \textbf{F1} & \textbf{Tok} & \textbf{Lat} \\
\midrule
\endfirsthead
\multicolumn{8}{c}{\tablename~\thetable{} -- continued}\\
\toprule
\textbf{Model} & \textbf{Input} & \textbf{Acc} & \textbf{Prec} & \textbf{Rec} & \textbf{F1} & \textbf{Tok} & \textbf{Lat} \\
\midrule
\endhead
\midrule
\multicolumn{8}{r}{\footnotesize continued on next page}\\
\endfoot
\bottomrule
\endlastfoot
      gemini-3-pro & asr\_diar\_meta & 0.89 & 0.50 & 0.82 & 0.62 & 6856 & 31932 \\
      gemini-3-pro & asr\_meta & 0.85 & 0.41 & 0.82 & 0.55 & 6732 & 29595 \\
      gemini-3-pro & meta & 0.88 & 0.47 & 0.64 & 0.54 & 6443 & 31008 \\
      gemini-3-pro & only & 0.88 & 0.47 & 0.73 & 0.57 & 6224 & 35729 \\
      molmo-2-8b & asr\_diar\_meta & 0.90 & 0.43 & 0.33 & 0.38 & 12197 & 22064 \\
      molmo-2-8b & asr\_meta & 0.84 & 0.27 & 0.44 & 0.33 & 11846 & 26757 \\
      molmo-2-8b & meta & 0.82 & 0.24 & 0.44 & 0.31 & 11608 & 24596 \\
      molmo-2-8b & only & 0.88 & 0.25 & 0.10 & 0.14 & 11361 & 21761 \\
      qwen3-omni & asr\_diar\_meta & 0.69 & 0.22 & 0.89 & 0.35 & 15346 & 9525 \\
      qwen3-omni & asr\_meta & 0.68 & 0.21 & 0.89 & 0.34 & 15207 & 9735 \\
      qwen3-omni & meta & 0.73 & 0.24 & 0.89 & 0.38 & 14861 & 9879 \\
      qwen3-omni & only & 0.91 & 0.57 & 0.40 & 0.47 & 14615 & 9913 \\
      qwen3-235b & asr\_diar\_meta & 0.77 & 0.23 & 0.67 & 0.34 & 14614 & 7178 \\
      qwen3-235b & asr\_meta & 0.74 & 0.19 & 0.63 & 0.29 & 14475 & 7237 \\
      qwen3-235b & meta & 0.79 & 0.23 & 0.56 & 0.32 & 14129 & 7091 \\
      qwen3-235b & only & 0.84 & 0.29 & 0.56 & 0.38 & 13884 & 9023 \\
      qwen3-30b & asr\_diar\_meta & 0.77 & 0.25 & 0.78 & 0.38 & 14614 & 22218 \\
      qwen3-30b & asr\_meta & 0.78 & 0.29 & 0.80 & 0.42 & 14475 & 13720 \\
      qwen3-30b & meta & 0.73 & 0.18 & 0.56 & 0.27 & 14130 & 20742 \\
      qwen3-30b & only & 0.71 & 0.17 & 0.63 & 0.26 & 13884 & 18405 \\
      seed-1.6 & asr\_diar\_meta & 0.88 & 0.33 & 0.09 & 0.14 & 19506 & 40369 \\
      seed-1.6 & asr\_meta & 0.89 & 0.50 & 0.36 & 0.42 & 19361 & 34226 \\
      seed-1.6 & meta & 0.87 & 0.00 & 0.00 & -- & 19036 & 33848 \\
      seed-1.6 & only & 0.89 & 1.00 & 0.08 & 0.15 & 18799 & 57734 \\
\end{longtable}
\end{center}

\begin{center}
\footnotesize
\setlength{\tabcolsep}{5pt}
\begin{longtable}{llrrrrrr}
\caption{Visual NER / Person Recognition --- video pipeline. Input: \texttt{asr\_diar\_meta}=video+ASR+diarization+metadata, \texttt{asr\_meta}=video+ASR+metadata, \texttt{meta}=video+metadata, \texttt{only}=video only. \textbf{Tok}=input tokens; \textbf{Lat}=API latency (ms); \texttt{--}=undefined (no positive predictions).}\label{tab:res_vid_per}\\
\toprule
\textbf{Model} & \textbf{Input} & \textbf{Acc} & \textbf{Prec} & \textbf{Rec} & \textbf{F1} & \textbf{Tok} & \textbf{Lat} \\
\midrule
\endfirsthead
\multicolumn{8}{c}{\tablename~\thetable{} -- continued}\\
\toprule
\textbf{Model} & \textbf{Input} & \textbf{Acc} & \textbf{Prec} & \textbf{Rec} & \textbf{F1} & \textbf{Tok} & \textbf{Lat} \\
\midrule
\endhead
\midrule
\multicolumn{8}{r}{\footnotesize continued on next page}\\
\endfoot
\bottomrule
\endlastfoot
      gemini-3-pro & asr\_diar\_meta & 0.60 & 0.62 & 0.60 & 0.61 & 6856 & 31932 \\
      gemini-3-pro & asr\_meta & 0.50 & 0.56 & 0.50 & 0.52 & 6732 & 29595 \\
      gemini-3-pro & meta & 0.60 & 0.63 & 0.60 & 0.60 & 6443 & 31008 \\
      gemini-3-pro & only & 0.50 & 0.54 & 0.50 & 0.50 & 6224 & 35729 \\
      molmo-2-8b & asr\_diar\_meta & 0.14 & 0.18 & 0.14 & 0.10 & 12197 & 22064 \\
      molmo-2-8b & asr\_meta & 0.13 & 0.11 & 0.13 & 0.07 & 11846 & 26757 \\
      molmo-2-8b & meta & 0.10 & 0.01 & 0.10 & 0.02 & 11608 & 24596 \\
      molmo-2-8b & only & 0.10 & 0.01 & 0.10 & 0.02 & 11361 & 21761 \\
      qwen3-omni & asr\_diar\_meta & 0.10 & 0.01 & 0.10 & 0.02 & 15346 & 9525 \\
      qwen3-omni & asr\_meta & 0.10 & 0.01 & 0.10 & 0.02 & 15207 & 9735 \\
      qwen3-omni & meta & 0.10 & 0.01 & 0.10 & 0.02 & 14861 & 9879 \\
      qwen3-omni & only & 0.10 & 0.01 & 0.10 & 0.02 & 14615 & 9913 \\
      qwen3-235b & asr\_diar\_meta & 0.13 & 0.14 & 0.13 & 0.06 & 14614 & 7178 \\
      qwen3-235b & asr\_meta & 0.18 & 0.17 & 0.18 & 0.12 & 14475 & 7237 \\
      qwen3-235b & meta & 0.13 & 0.09 & 0.13 & 0.06 & 14129 & 7091 \\
      qwen3-235b & only & 0.15 & 0.06 & 0.15 & 0.07 & 13884 & 9023 \\
      qwen3-30b & asr\_diar\_meta & 0.10 & 0.01 & 0.10 & 0.02 & 14614 & 22218 \\
      qwen3-30b & asr\_meta & 0.10 & 0.01 & 0.10 & 0.02 & 14475 & 13720 \\
      qwen3-30b & meta & 0.10 & 0.01 & 0.10 & 0.02 & 14130 & 20742 \\
      qwen3-30b & only & 0.18 & 0.10 & 0.18 & 0.12 & 13884 & 18405 \\
      seed-1.6 & asr\_diar\_meta & 0.42 & 0.40 & 0.42 & 0.38 & 19506 & 40369 \\
      seed-1.6 & asr\_meta & 0.39 & 0.38 & 0.39 & 0.37 & 19361 & 34226 \\
      seed-1.6 & meta & 0.31 & 0.31 & 0.31 & 0.26 & 19036 & 33848 \\
      seed-1.6 & only & 0.10 & 0.01 & 0.10 & 0.02 & 18799 & 57734 \\
\end{longtable}
\end{center}

\begin{center}
\footnotesize
\setlength{\tabcolsep}{5pt}
\begin{longtable}{llrrrrrr}
\caption{Environment Recognition --- frame pipeline. Input: \texttt{asr}=frames+ASR, \texttt{asr\_diar}=+diarization, \texttt{asr\_diar\_meta}=+diarization+metadata, \texttt{asr\_meta}=frames+ASR+metadata, \texttt{only}=frames only. Columns as in Table~\ref{tab:res_vid_env}.}\label{tab:res_frm_env}\\
\toprule
\textbf{Model} & \textbf{Input} & \textbf{Acc} & \textbf{Prec} & \textbf{Rec} & \textbf{F1} & \textbf{Tok} & \textbf{Lat} \\
\midrule
\endfirsthead
\multicolumn{8}{c}{\tablename~\thetable{} -- continued}\\
\toprule
\textbf{Model} & \textbf{Input} & \textbf{Acc} & \textbf{Prec} & \textbf{Rec} & \textbf{F1} & \textbf{Tok} & \textbf{Lat} \\
\midrule
\endhead
\midrule
\multicolumn{8}{r}{\footnotesize continued on next page}\\
\endfoot
\bottomrule
\endlastfoot
      gemini-3-pro & asr & 0.68 & 0.72 & 0.68 & 0.68 & 14294 & 35028 \\
      gemini-3-pro & asr\_diar & 0.72 & 0.77 & 0.72 & 0.72 & 14456 & 33811 \\
      gemini-3-pro & asr\_diar\_meta & 0.73 & 0.79 & 0.73 & 0.74 & 14666 & 34378 \\
      gemini-3-pro & asr\_meta & 0.76 & 0.81 & 0.76 & 0.77 & 14572 & 32011 \\
      gemini-3-pro & only & 0.70 & 0.75 & 0.70 & 0.71 & 13996 & 34865 \\
      gemma-3-27b & asr & 0.46 & 0.51 & 0.46 & 0.43 & 4212 & 25520 \\
      gemma-3-27b & asr\_diar & 0.47 & 0.29 & 0.47 & 0.36 & 4376 & 28385 \\
      gemma-3-27b & asr\_diar\_meta & 0.44 & 0.30 & 0.44 & 0.34 & 4586 & 8738 \\
      gemma-3-27b & asr\_meta & 0.37 & 0.47 & 0.37 & 0.34 & 4491 & 29150 \\
      gemma-3-27b & only & 0.49 & 0.45 & 0.49 & 0.45 & 3917 & 8167 \\
      gpt-5-mini & asr & 0.55 & 0.50 & 0.55 & 0.52 & 4551 & 37385 \\
      gpt-5-mini & asr\_diar & 0.27 & 0.51 & 0.27 & 0.34 & 4705 & 55330 \\
      gpt-5-mini & asr\_diar\_meta & 0.56 & 0.62 & 0.56 & 0.58 & 4945 & 28378 \\
      gpt-5-mini & asr\_meta & 0.46 & 0.72 & 0.46 & 0.54 & 4849 & 45219 \\
      gpt-5-mini & only & 0.59 & 0.64 & 0.59 & 0.59 & 4234 & 15836 \\
      llama-4-mav & asr & 0.57 & 0.64 & 0.57 & 0.59 & 9753 & 20462 \\
      llama-4-mav & asr\_diar & 0.58 & 0.56 & 0.58 & 0.54 & 9906 & 24420 \\
      llama-4-mav & asr\_diar\_meta & 0.60 & 0.68 & 0.60 & 0.57 & 10135 & 8195 \\
      llama-4-mav & asr\_meta & 0.54 & 0.62 & 0.54 & 0.56 & 10056 & 19666 \\
      llama-4-mav & only & 0.55 & 0.60 & 0.55 & 0.56 & 9461 & 4710 \\
      ministral-14b & asr & 0.28 & 0.30 & 0.28 & 0.25 & 5441 & 19854 \\
      ministral-14b & asr\_diar & 0.21 & 0.37 & 0.21 & 0.15 & 5620 & 21727 \\
      ministral-14b & asr\_diar\_meta & 0.20 & 0.35 & 0.20 & 0.22 & 5868 & 10670 \\
      ministral-14b & asr\_meta & 0.08 & 0.22 & 0.08 & 0.10 & 5742 & 26039 \\
      ministral-14b & only & 0.25 & 0.27 & 0.25 & 0.22 & 5124 & 7118 \\
      qwen2.5-72b & asr & 0.60 & 0.62 & 0.60 & 0.58 & 4821 & 19667 \\
      qwen2.5-72b & asr\_diar & 0.53 & 0.47 & 0.53 & 0.46 & 4983 & 23429 \\
      qwen2.5-72b & asr\_diar\_meta & 0.59 & 0.47 & 0.59 & 0.50 & 5241 & 6494 \\
      qwen2.5-72b & asr\_meta & 0.56 & 0.51 & 0.56 & 0.52 & 5150 & 19650 \\
      qwen2.5-72b & only & 0.63 & 0.67 & 0.63 & 0.62 & 4460 & 7222 \\
      qwen3-235b & asr & 0.65 & 0.63 & 0.65 & 0.62 & 3863 & 21662 \\
      qwen3-235b & asr\_diar & 0.57 & 0.58 & 0.57 & 0.54 & 4024 & 21450 \\
      qwen3-235b & asr\_diar\_meta & 0.62 & 0.58 & 0.62 & 0.57 & 4283 & 6062 \\
      qwen3-235b & asr\_meta & 0.67 & 0.70 & 0.67 & 0.67 & 4192 & 20654 \\
      qwen3-235b & only & 0.65 & 0.62 & 0.65 & 0.63 & 3501 & 3617 \\
      qwen3-30b & asr & 0.56 & 0.55 & 0.56 & 0.53 & 3863 & 19758 \\
      qwen3-30b & asr\_diar & 0.44 & 0.36 & 0.44 & 0.35 & 4024 & 16779 \\
      qwen3-30b & asr\_diar\_meta & 0.41 & 0.41 & 0.41 & 0.34 & 4283 & 5855 \\
      qwen3-30b & asr\_meta & 0.30 & 0.49 & 0.30 & 0.28 & 4192 & 16132 \\
      qwen3-30b & only & 0.50 & 0.52 & 0.50 & 0.47 & 3501 & 4657 \\
      qwen3-8b & asr & 0.34 & 0.53 & 0.34 & 0.32 & 9308 & 24031 \\
      qwen3-8b & asr\_diar & 0.43 & 0.34 & 0.43 & 0.34 & 7690 & 18707 \\
      qwen3-8b & asr\_diar\_meta & 0.43 & 0.36 & 0.43 & 0.37 & 4283 & 7789 \\
      qwen3-8b & asr\_meta & 0.36 & 0.40 & 0.36 & 0.33 & 7835 & 23203 \\
      qwen3-8b & only & 0.49 & 0.61 & 0.49 & 0.48 & 3501 & 7308 \\
\end{longtable}
\end{center}

\begin{center}
\footnotesize
\setlength{\tabcolsep}{5pt}
\begin{longtable}{llrrrrrr}
\caption{Topic Classification --- frame pipeline. Input: \texttt{asr}=frames+ASR, \texttt{asr\_diar}=+diarization, \texttt{asr\_diar\_meta}=+diarization+metadata, \texttt{asr\_meta}=frames+ASR+metadata, \texttt{only}=frames only. Columns as in Table~\ref{tab:res_vid_env}.}\label{tab:res_frm_top}\\
\toprule
\textbf{Model} & \textbf{Input} & \textbf{Acc} & \textbf{Prec} & \textbf{Rec} & \textbf{F1} & \textbf{Tok} & \textbf{Lat} \\
\midrule
\endfirsthead
\multicolumn{8}{c}{\tablename~\thetable{} -- continued}\\
\toprule
\textbf{Model} & \textbf{Input} & \textbf{Acc} & \textbf{Prec} & \textbf{Rec} & \textbf{F1} & \textbf{Tok} & \textbf{Lat} \\
\midrule
\endhead
\midrule
\multicolumn{8}{r}{\footnotesize continued on next page}\\
\endfoot
\bottomrule
\endlastfoot
      gemini-3-pro & asr & 0.81 & 0.82 & 0.81 & 0.80 & 14294 & 35028 \\
      gemini-3-pro & asr\_diar & 0.81 & 0.84 & 0.81 & 0.80 & 14456 & 33811 \\
      gemini-3-pro & asr\_diar\_meta & 0.80 & 0.81 & 0.80 & 0.78 & 14666 & 34378 \\
      gemini-3-pro & asr\_meta & 0.79 & 0.81 & 0.79 & 0.78 & 14572 & 32011 \\
      gemini-3-pro & only & 0.67 & 0.73 & 0.67 & 0.65 & 13996 & 34865 \\
      gemma-3-27b & asr & 0.65 & 0.72 & 0.65 & 0.67 & 4212 & 25520 \\
      gemma-3-27b & asr\_diar & 0.68 & 0.75 & 0.68 & 0.69 & 4376 & 28385 \\
      gemma-3-27b & asr\_diar\_meta & 0.67 & 0.73 & 0.67 & 0.68 & 4586 & 8738 \\
      gemma-3-27b & asr\_meta & 0.65 & 0.74 & 0.65 & 0.66 & 4491 & 29150 \\
      gemma-3-27b & only & 0.50 & 0.62 & 0.50 & 0.50 & 3917 & 8167 \\
      gpt-5-mini & asr & 0.70 & 0.71 & 0.70 & 0.69 & 4551 & 37385 \\
      gpt-5-mini & asr\_diar & 0.39 & 0.74 & 0.39 & 0.49 & 4705 & 55330 \\
      gpt-5-mini & asr\_diar\_meta & 0.63 & 0.80 & 0.63 & 0.68 & 4945 & 28378 \\
      gpt-5-mini & asr\_meta & 0.47 & 0.75 & 0.47 & 0.55 & 4849 & 45219 \\
      gpt-5-mini & only & 0.61 & 0.73 & 0.61 & 0.63 & 4234 & 15836 \\
      llama-4-mav & asr & 0.71 & 0.76 & 0.71 & 0.71 & 9753 & 20462 \\
      llama-4-mav & asr\_diar & 0.71 & 0.76 & 0.71 & 0.70 & 9906 & 24420 \\
      llama-4-mav & asr\_diar\_meta & 0.69 & 0.72 & 0.69 & 0.67 & 10135 & 8195 \\
      llama-4-mav & asr\_meta & 0.71 & 0.75 & 0.71 & 0.71 & 10056 & 19666 \\
      llama-4-mav & only & 0.50 & 0.65 & 0.50 & 0.52 & 9461 & 4710 \\
      ministral-14b & asr & 0.53 & 0.66 & 0.53 & 0.54 & 5441 & 19854 \\
      ministral-14b & asr\_diar & 0.53 & 0.69 & 0.53 & 0.57 & 5620 & 21727 \\
      ministral-14b & asr\_diar\_meta & 0.35 & 0.71 & 0.35 & 0.46 & 5868 & 10670 \\
      ministral-14b & asr\_meta & 0.13 & 0.54 & 0.13 & 0.20 & 5742 & 26039 \\
      ministral-14b & only & 0.26 & 0.47 & 0.26 & 0.31 & 5124 & 7118 \\
      qwen2.5-72b & asr & 0.69 & 0.84 & 0.69 & 0.72 & 4821 & 19667 \\
      qwen2.5-72b & asr\_diar & 0.63 & 0.78 & 0.63 & 0.66 & 4983 & 23429 \\
      qwen2.5-72b & asr\_diar\_meta & 0.66 & 0.75 & 0.66 & 0.67 & 5241 & 6494 \\
      qwen2.5-72b & asr\_meta & 0.74 & 0.78 & 0.74 & 0.74 & 5150 & 19650 \\
      qwen2.5-72b & only & 0.53 & 0.70 & 0.53 & 0.55 & 4460 & 7222 \\
      qwen3-235b & asr & 0.66 & 0.75 & 0.66 & 0.68 & 3863 & 21662 \\
      qwen3-235b & asr\_diar & 0.65 & 0.76 & 0.65 & 0.67 & 4024 & 21450 \\
      qwen3-235b & asr\_diar\_meta & 0.65 & 0.72 & 0.65 & 0.64 & 4283 & 6062 \\
      qwen3-235b & asr\_meta & 0.70 & 0.72 & 0.70 & 0.69 & 4192 & 20654 \\
      qwen3-235b & only & 0.52 & 0.66 & 0.52 & 0.55 & 3501 & 3617 \\
      qwen3-30b & asr & 0.66 & 0.78 & 0.66 & 0.68 & 3863 & 19758 \\
      qwen3-30b & asr\_diar & 0.59 & 0.75 & 0.59 & 0.61 & 4024 & 16779 \\
      qwen3-30b & asr\_diar\_meta & 0.62 & 0.68 & 0.62 & 0.61 & 4283 & 5855 \\
      qwen3-30b & asr\_meta & 0.67 & 0.76 & 0.67 & 0.67 & 4192 & 16132 \\
      qwen3-30b & only & 0.47 & 0.69 & 0.47 & 0.49 & 3501 & 4657 \\
      qwen3-8b & asr & 0.50 & 0.67 & 0.50 & 0.52 & 9308 & 24031 \\
      qwen3-8b & asr\_diar & 0.49 & 0.63 & 0.49 & 0.49 & 7690 & 18707 \\
      qwen3-8b & asr\_diar\_meta & 0.52 & 0.69 & 0.52 & 0.51 & 4283 & 7789 \\
      qwen3-8b & asr\_meta & 0.52 & 0.60 & 0.52 & 0.52 & 7835 & 23203 \\
      qwen3-8b & only & 0.39 & 0.62 & 0.39 & 0.38 & 3501 & 7308 \\
\end{longtable}
\end{center}

\begin{center}
\footnotesize
\setlength{\tabcolsep}{5pt}
\begin{longtable}{llrrrrrr}
\caption{Sensitive-Content Detection --- frame pipeline. \emph{Metrics are diagnostic only}: with 12 positive clips, precision/recall are highly unstable and are \emph{not} used for configuration selection. Input: \texttt{asr}=frames+ASR, \texttt{asr\_diar}=+diarization, \texttt{asr\_diar\_meta}=+diarization+metadata, \texttt{asr\_meta}=frames+ASR+metadata, \texttt{only}=frames only. Columns as in Table~\ref{tab:res_vid_env}.}\label{tab:res_frm_sen}\\
\toprule
\textbf{Model} & \textbf{Input} & \textbf{Acc} & \textbf{Prec} & \textbf{Rec} & \textbf{F1} & \textbf{Tok} & \textbf{Lat} \\
\midrule
\endfirsthead
\multicolumn{8}{c}{\tablename~\thetable{} -- continued}\\
\toprule
\textbf{Model} & \textbf{Input} & \textbf{Acc} & \textbf{Prec} & \textbf{Rec} & \textbf{F1} & \textbf{Tok} & \textbf{Lat} \\
\midrule
\endhead
\midrule
\multicolumn{8}{r}{\footnotesize continued on next page}\\
\endfoot
\bottomrule
\endlastfoot
      gemini-3-pro & asr & 0.83 & 0.38 & 0.82 & 0.51 & 14294 & 35028 \\
      gemini-3-pro & asr\_diar & 0.86 & 0.43 & 0.82 & 0.56 & 14456 & 33811 \\
      gemini-3-pro & asr\_diar\_meta & 0.77 & 0.27 & 0.64 & 0.38 & 14666 & 34378 \\
      gemini-3-pro & asr\_meta & 0.86 & 0.40 & 0.80 & 0.53 & 14572 & 32011 \\
      gemini-3-pro & only & 0.87 & 0.43 & 0.55 & 0.48 & 13996 & 34865 \\
      gemma-3-27b & asr & 0.87 & 0.40 & 0.60 & 0.48 & 4212 & 25520 \\
      gemma-3-27b & asr\_diar & 0.89 & 0.40 & 0.20 & 0.27 & 4376 & 28385 \\
      gemma-3-27b & asr\_diar\_meta & 0.43 & 0.05 & 0.43 & 0.10 & 4586 & 8738 \\
      gemma-3-27b & asr\_meta & 0.89 & 0.43 & 0.30 & 0.35 & 4491 & 29150 \\
      gemma-3-27b & only & 0.90 & 0.44 & 0.44 & 0.44 & 3917 & 8167 \\
      gpt-5-mini & asr & 0.70 & 0.22 & 0.89 & 0.35 & 4551 & 37385 \\
      gpt-5-mini & asr\_diar & 0.84 & 0.20 & 0.20 & 0.20 & 4705 & 55330 \\
      gpt-5-mini & asr\_diar\_meta & 0.76 & 0.23 & 0.60 & 0.33 & 4945 & 28378 \\
      gpt-5-mini & asr\_meta & 0.85 & 0.36 & 0.33 & 0.35 & 4849 & 45219 \\
      gpt-5-mini & only & 0.81 & 0.25 & 0.36 & 0.30 & 4234 & 15836 \\
      llama-4-mav & asr & 0.65 & 0.17 & 0.88 & 0.29 & 9753 & 20462 \\
      llama-4-mav & asr\_diar & 0.91 & 0.57 & 0.40 & 0.47 & 9906 & 24420 \\
      llama-4-mav & asr\_diar\_meta & 0.13 & 0.00 & -- & -- & 10135 & 8195 \\
      llama-4-mav & asr\_meta & 0.33 & 0.00 & 0.00 & -- & 10056 & 19666 \\
      llama-4-mav & only & 0.72 & 0.19 & 0.67 & 0.30 & 9461 & 4710 \\
      ministral-14b & asr & 0.88 & -- & 0.00 & -- & 5441 & 19854 \\
      ministral-14b & asr\_diar & 0.88 & -- & 0.00 & -- & 5620 & 21727 \\
      ministral-14b & asr\_diar\_meta & 0.88 & -- & 0.00 & -- & 5868 & 10670 \\
      ministral-14b & asr\_meta & 0.87 & 0.00 & 0.00 & -- & 5742 & 26039 \\
      ministral-14b & only & 0.88 & -- & 0.00 & -- & 5124 & 7118 \\
      qwen2.5-72b & asr & 0.87 & 0.43 & 0.25 & 0.32 & 4821 & 19667 \\
      qwen2.5-72b & asr\_diar & 0.84 & 0.25 & 0.30 & 0.27 & 4983 & 23429 \\
      qwen2.5-72b & asr\_diar\_meta & 0.88 & 0.00 & 0.00 & -- & 5241 & 6494 \\
      qwen2.5-72b & asr\_meta & 0.89 & 0.50 & 0.09 & 0.15 & 5150 & 19650 \\
      qwen2.5-72b & only & 0.91 & 0.80 & 0.33 & 0.47 & 4460 & 7222 \\
      qwen3-235b & asr & 0.84 & 0.14 & 0.33 & 0.20 & 3863 & 21662 \\
      qwen3-235b & asr\_diar & 0.82 & 0.26 & 0.86 & 0.40 & 4024 & 21450 \\
      qwen3-235b & asr\_diar\_meta & 0.84 & 0.15 & 0.29 & 0.20 & 4283 & 6062 \\
      qwen3-235b & asr\_meta & 0.81 & 0.12 & 0.33 & 0.17 & 4192 & 20654 \\
      qwen3-235b & only & 0.89 & 0.40 & 0.20 & 0.27 & 3501 & 3617 \\
      qwen3-30b & asr & 0.82 & 0.32 & 0.70 & 0.44 & 3863 & 19758 \\
      qwen3-30b & asr\_diar & 0.63 & 0.20 & 1.00 & 0.33 & 4024 & 16779 \\
      qwen3-30b & asr\_diar\_meta & 0.86 & 0.25 & 0.20 & 0.22 & 4283 & 5855 \\
      qwen3-30b & asr\_meta & 0.88 & 0.41 & 0.78 & 0.54 & 4192 & 16132 \\
      qwen3-30b & only & 0.81 & 0.13 & 0.29 & 0.17 & 3501 & 4657 \\
      qwen3-8b & asr & 0.86 & 0.25 & 0.20 & 0.22 & 9308 & 24031 \\
      qwen3-8b & asr\_diar & 0.80 & 0.17 & 0.38 & 0.23 & 7690 & 18707 \\
      qwen3-8b & asr\_diar\_meta & 0.01 & 0.01 & 1.00 & 0.02 & 4283 & 7789 \\
      qwen3-8b & asr\_meta & 0.19 & 0.02 & 1.00 & 0.05 & 7835 & 23203 \\
      qwen3-8b & only & 0.87 & 0.40 & 0.17 & 0.24 & 3501 & 7308 \\
\end{longtable}
\end{center}

\begin{center}
\footnotesize
\setlength{\tabcolsep}{5pt}
\begin{longtable}{llrrrrrr}
\caption{Visual NER / Person Recognition --- frame pipeline. Input: \texttt{asr}=frames+ASR, \texttt{asr\_diar}=+diarization, \texttt{asr\_diar\_meta}=+diarization+metadata, \texttt{asr\_meta}=frames+ASR+metadata, \texttt{only}=frames only. Columns as in Table~\ref{tab:res_vid_env}.}\label{tab:res_frm_per}\\
\toprule
\textbf{Model} & \textbf{Input} & \textbf{Acc} & \textbf{Prec} & \textbf{Rec} & \textbf{F1} & \textbf{Tok} & \textbf{Lat} \\
\midrule
\endfirsthead
\multicolumn{8}{c}{\tablename~\thetable{} -- continued}\\
\toprule
\textbf{Model} & \textbf{Input} & \textbf{Acc} & \textbf{Prec} & \textbf{Rec} & \textbf{F1} & \textbf{Tok} & \textbf{Lat} \\
\midrule
\endhead
\midrule
\multicolumn{8}{r}{\footnotesize continued on next page}\\
\endfoot
\bottomrule
\endlastfoot
      gemini-3-pro & asr & 0.39 & 0.42 & 0.39 & 0.38 & 14294 & 35028 \\
      gemini-3-pro & asr\_diar & 0.27 & 0.29 & 0.27 & 0.24 & 14456 & 33811 \\
      gemini-3-pro & asr\_diar\_meta & 0.30 & 0.27 & 0.30 & 0.26 & 14666 & 34378 \\
      gemini-3-pro & asr\_meta & 0.59 & 0.60 & 0.59 & 0.58 & 14572 & 32011 \\
      gemini-3-pro & only & 0.43 & 0.39 & 0.43 & 0.40 & 13996 & 34865 \\
      gemma-3-27b & asr & 0.15 & 0.08 & 0.15 & 0.09 & 4212 & 25520 \\
      gemma-3-27b & asr\_diar & 0.16 & 0.09 & 0.16 & 0.10 & 4376 & 28385 \\
      gemma-3-27b & asr\_diar\_meta & 0.20 & 0.14 & 0.20 & 0.16 & 4586 & 8738 \\
      gemma-3-27b & asr\_meta & 0.36 & 0.38 & 0.36 & 0.36 & 4491 & 29150 \\
      gemma-3-27b & only & 0.15 & 0.06 & 0.15 & 0.07 & 3917 & 8167 \\
      gpt-5-mini & asr & 0.09 & 0.06 & 0.09 & 0.07 & 4551 & 37385 \\
      gpt-5-mini & asr\_diar & 0.11 & 0.05 & 0.11 & 0.06 & 4705 & 55330 \\
      gpt-5-mini & asr\_diar\_meta & 0.21 & 0.26 & 0.21 & 0.20 & 4945 & 28378 \\
      gpt-5-mini & asr\_meta & 0.11 & 0.04 & 0.11 & 0.04 & 4849 & 45219 \\
      gpt-5-mini & only & 0.10 & 0.01 & 0.10 & 0.02 & 4234 & 15836 \\
      llama-4-mav & asr & 0.13 & 0.07 & 0.13 & 0.08 & 9753 & 20462 \\
      llama-4-mav & asr\_diar & 0.12 & 0.10 & 0.12 & 0.09 & 9906 & 24420 \\
      llama-4-mav & asr\_diar\_meta & 0.26 & 0.36 & 0.26 & 0.28 & 10135 & 8195 \\
      llama-4-mav & asr\_meta & 0.11 & 0.05 & 0.11 & 0.04 & 10056 & 19666 \\
      llama-4-mav & only & 0.16 & 0.08 & 0.16 & 0.09 & 9461 & 4710 \\
      ministral-14b & asr & 0.09 & 0.01 & 0.09 & 0.02 & 5441 & 19854 \\
      ministral-14b & asr\_diar & 0.09 & 0.02 & 0.09 & 0.03 & 5620 & 21727 \\
      ministral-14b & asr\_diar\_meta & 0.14 & 0.09 & 0.14 & 0.09 & 5868 & 10670 \\
      ministral-14b & asr\_meta & 0.14 & 0.12 & 0.14 & 0.08 & 5742 & 26039 \\
      ministral-14b & only & 0.09 & 0.01 & 0.09 & 0.02 & 5124 & 7118 \\
      qwen2.5-72b & asr & 0.13 & 0.05 & 0.13 & 0.06 & 4821 & 19667 \\
      qwen2.5-72b & asr\_diar & 0.12 & 0.05 & 0.12 & 0.06 & 4983 & 23429 \\
      qwen2.5-72b & asr\_diar\_meta & 0.21 & 0.25 & 0.21 & 0.20 & 5241 & 6494 \\
      qwen2.5-72b & asr\_meta & 0.48 & 0.45 & 0.48 & 0.45 & 5150 & 19650 \\
      qwen2.5-72b & only & 0.13 & 0.04 & 0.13 & 0.05 & 4460 & 7222 \\
      qwen3-235b & asr & 0.17 & 0.08 & 0.17 & 0.10 & 3863 & 21662 \\
      qwen3-235b & asr\_diar & 0.13 & 0.07 & 0.13 & 0.08 & 4024 & 21450 \\
      qwen3-235b & asr\_diar\_meta & 0.22 & 0.16 & 0.22 & 0.17 & 4283 & 6062 \\
      qwen3-235b & asr\_meta & 0.48 & 0.52 & 0.48 & 0.49 & 4192 & 20654 \\
      qwen3-235b & only & 0.15 & 0.06 & 0.15 & 0.07 & 3501 & 3617 \\
      qwen3-30b & asr & 0.15 & 0.08 & 0.15 & 0.10 & 3863 & 19758 \\
      qwen3-30b & asr\_diar & 0.13 & 0.08 & 0.13 & 0.09 & 4024 & 16779 \\
      qwen3-30b & asr\_diar\_meta & 0.23 & 0.29 & 0.23 & 0.23 & 4283 & 5855 \\
      qwen3-30b & asr\_meta & 0.41 & 0.52 & 0.41 & 0.43 & 4192 & 16132 \\
      qwen3-30b & only & 0.12 & 0.04 & 0.12 & 0.05 & 3501 & 4657 \\
      qwen3-8b & asr & 0.14 & 0.06 & 0.14 & 0.07 & 9308 & 24031 \\
      qwen3-8b & asr\_diar & 0.14 & 0.07 & 0.14 & 0.08 & 7690 & 18707 \\
      qwen3-8b & asr\_diar\_meta & 0.24 & 0.27 & 0.24 & 0.25 & 4283 & 7789 \\
      qwen3-8b & asr\_meta & 0.29 & 0.30 & 0.29 & 0.28 & 7835 & 23203 \\
      qwen3-8b & only & 0.11 & 0.03 & 0.11 & 0.04 & 3501 & 7308 \\
\end{longtable}
\end{center}
\end{document}